\documentclass[11pt]{article}
\usepackage{fullname}
\usepackage{epsfig}
\usepackage{hhline}

\advance\textheight by \topskip
\newcommand{\pat}[2]{{#1$_{#2}$}}
\newcommand{\slot}[3]{{#1$_{#2}\!=\,$#3}}
\newcommand{\nleft}{{[$_N$}}
\newcommand{\nright}{{$_N$]}}
\newcommand{\vleft}{{[$_V$}}
\newcommand{\vright}{{$_V$]}}
\newcommand{\tag}[1]{{\tt\large #1}}
\newcommand{\tagi}[1]{{\tt\large \hspace*{0.75em} #1}}

\begin{document}
\author{Lance A. Ramshaw\\
Dept. of Computer Science\\
Bowdoin College\\
Brunswick, ME  04011 USA\\
{\tt ramshaw@polar.bowdoin.edu} \and
Mitchell P. Marcus\\
Computer and Information Science Dept.\\
University of Pennsylvania\\
%558 Moore Building\\
Philadelphia, PA  19104--6389 USA\\
{\tt mitch@linc.cis.upenn.edu}
}
\title{Text Chunking using Transformation-Based Learning}
\date{}
\maketitle

\begin{abstract}
Eric Brill introduced
transformation-based learning
and showed that it can do part-of-speech tagging
with fairly high accuracy.
The same method can be applied
at a higher level of textual interpretation
for locating chunks in the tagged text,
including non-recursive ``baseNP'' chunks.
For this purpose, it is convenient
to view chunking as a tagging problem
by encoding the chunk structure in new tags attached to each word.
In automatic tests using Treebank-derived data,
this technique achieved recall and precision rates of roughly 92\%
for baseNP chunks and 88\%
for somewhat more complex chunks that partition the sentence.
Some interesting adaptations
to the transformation-based learning approach
are also suggested by this application.
\end{abstract}

\bibliographystyle{fullname}

\section{Introduction}

Text chunking involves dividing sentences into
nonoverlapping segments
on the basis of fairly superficial analysis.
Abney \shortcite{Abney91} has proposed this
as a useful and relatively tractable precursor
to full parsing,
since it provides a foundation
for further levels of analysis including verb-argument identification,
while still allowing more complex attachment decisions
to be postponed to a later phase.
Since chunking includes identifying
the non-recursive portions of noun phrases,
it can also be useful for other purposes
including index term generation.

Most efforts at superficially extracting segments from sentences
have focused on identifying low-level noun groups, either
using hand-built grammars and finite state techniques or
using statistical models like HMMs trained from corpora.
In this paper, we target a somewhat higher level of chunk structure
using Brill's \shortcite{Brill93}
transformation-based learning mechanism,
in which a sequence of transformational rules
is learned from a corpus;
this sequence iteratively improves upon a baseline model
for some interpretive feature of the text.
This technique has previously been used not only for
part-of-speech tagging \cite{Brill94b}, but also for
prepositional phrase attachment disambiguation \cite{BrillResnik94},
% identifying verb-argument structure,
and assigning unlabeled binary-branching tree structure
to sentences \cite{Brill93a}.
Because transformation-based learning uses
pattern-action rules based on selected features of the local context,
it is helpful for the values being predicted
to also be encoded locally.
In the text-chunking application,
encoding the predicted chunk structure
in tags attached to the words,
rather than as brackets between words,
avoids many of the difficulties with unbalanced bracketings
that would result if such local rules were allowed
to insert or alter inter-word brackets directly.

In this study,
training and test sets marked
with two different types of chunk structure
were derived algorithmically from the parsed data
in the Penn Treebank corpus
of Wall Street Journal text \cite{Marcus:treebank}.
The source texts were then run through Brill's
part-of-speech tagger \cite{Brill:tagger},
and, as a baseline heuristic,
chunk structure tags were assigned to each word
based on its part-of-speech tag.
Rules were then automatically learned
that updated these chunk structure tags
based on neighboring words
and their part-of-speech and chunk tags.
Applying transformation-based learning to text chunking
turns out to be different in interesting ways
from its use for part-of-speech tagging.
The much smaller tagset calls for a different organization
of the computation, and the fact that
part-of-speech assignments as well as word identities are fixed
suggests different optimizations.

\section{Text Chunking}

Abney \shortcite{Abney91} has proposed text chunking
as a useful preliminary step to parsing.
His chunks are inspired in part by psychological studies
of Gee and Grosjean \shortcite{GeeGrosjean83}
that link pause durations in reading
and naive sentence diagraming to
text groupings that they called $\phi$-phrases,
which very roughly correspond
to breaking the string after each syntactic
head that is a content word.
Abney's other motivation for chunking
is procedural, based on the hypothesis that the identification of chunks
can be done fairly dependably by finite state methods,
postponing the decisions that require higher-level analysis
to a parsing phase that chooses how to combine the chunks.

\subsection{Existing Chunk Identification Techniques}

Existing efforts at identifying chunks in text have been focused
primarily
on low-level noun group identification,
frequently as a step in deriving index terms,
motivated in part by the limited coverage of present broad-scale parsers
when dealing with unrestricted text.
Some researchers have applied grammar-based methods,
combining lexical data with finite state or other grammar constraints,
while others have worked on inducing statistical models
either directly from the words or from automatically assigned
part-of-speech classes.

On the grammar-based side,
Bourigault \shortcite{Bourigault92}
describes a system for extracting
``terminological noun phrases'' from French text.
This system first uses heuristics
to find ``maximal length noun phrases'',
and then uses a grammar to extract
``terminological units.''
For example, from the maximal NP
{\em le disque dur de la station de travail}
it extracts the two terminological phrases
{\em disque dur,} and {\em station de travail.}
Bourigault claims that the grammar can parse
``around 95\% of the maximal length noun phrases''
in a test corpus into possible terminological phrases,
which then require manual validation.
However, because its goal is terminological phrases,
it appears that this system ignores NP chunk-initial
determiners and
other initial prenominal modifiers,
somewhat simplifying the parsing task.

Voutilainen \shortcite{Voutilainen93},
in his impressive NPtool system,
uses an approach that is in some ways similar
to the one used here,
in that he adds to his part-of-speech tags
a new kind of tag that shows chunk structure;
the chunk tag ``@$>$N'', for example,
is used for determiners and premodifiers,
both of which group with the following noun head.
He uses a lexicon that lists all the possible chunk tags
for each word
combined with hand-built constraint grammar patterns.
These patterns eliminate impossible readings
to identify a somewhat idiosyncratic kind of target noun group
that does not include initial determiners but does include
postmodifying prepositional phrases (including determiners).
Voutilainen claims recall rates of 98.5\% or better with
precision of 95\% or better.
However, the sample NPtool analysis
given in the appendix of \cite{Voutilainen93},
appears to be less accurate than claimed in general,
with 5 apparent mistakes (and one unresolved ambiguity) out of the 32 NP
chunks in that sample, as listed in
Table~\ref{NPtool:errors}. These putative errors, combined with the
claimed high performance, suggest that NPtool's definition of NP chunk
is also tuned for extracting terminological phrases, and thus
excludes many kinds of NP premodifiers, again simplifying the chunking
task.
\begin{table}[th!]
\centering
\begin{tabular}{|c|c|} \hline
{\em NPtool parse} 	& {\em Apparent correct parse} \\ \hhline{|=|=|}
% for old-LaTeX, use \hline \hline in place of \hhline{}
less [time] 	& [less time]\\ \hline
the other hand 	& the [other hand]\\ \hline
 many [advantages] 	& [many advantages]\\ \hline
[binary addressing]  		& [binary addressing and \\
and [instruction formats]	& instruction formats] \\ \hline
a purely [binary computer] 	& a [purely binary computer] \\ \hline
\end{tabular}
\caption{Apparent errors made by Voutilainen's NPtool}
\label{NPtool:errors}
\end{table}

Kupiec \shortcite{Kupiec93} also briefly mentions the use of
finite state NP recognizers for both English and French
to prepare the input for a program that identified
the correspondences between NPs in bilingual corpora,
but he does not directly discuss their performance.

Using statistical methods,
Church's Parts program \shortcite{Church88},
in addition to identifying parts of speech,
also inserted brackets identifying core NPs.
These brackets were placed using a statistical model
trained on Brown corpus
material in which NP brackets had been inserted
semi-automatically.
In the small test sample shown,
this system achieved 98\% recall for correct brackets.
At about the same time, Ejerhed \shortcite{Ejerhed88},
working with Church, performed comparisons
between finite state methods and Church's stochastic models
for identifying both non-recursive clauses
and non-recursive NPs in English text.
In those comparisons, the stochastic methods
outperformed the hand built finite-state models,
with claimed accuracies of 93.5\% (clauses)
and 98.6\% (NPs) for the statistical models
compared to
to 87\% (clauses) and 97.8\% (NPs) for
the finite-state methods.

Running Church's program on test material,
however, reveals that the definition of NP
embodied in Church's program is quite simplified
in that it does not include, for example,
structures or words conjoined within NP by either explicit conjunctions
like ``and'' and ``or'', or implicitly by commas.
Church's chunker thus assigns
the following NP chunk structures:

%The {}'s are required to keep LateX from misinterpreting the []s!!a
\begin{center}
[a Skokie] , [Ill.] , [subsidiary] \\
{[newer]} , [big-selling prescriptions drugs] \\
{[the inefficiency]} , [waste] and [lack] of [coordination]\\
{[Kidder]} , [Peabody] \&  [Co]
\end{center}

It is difficult to compare performance figures between studies;
the definitions of the target chunks
and the evaluation methodologies differ widely
and are frequently incompletely specified.
All of the cited performance figures above also appear to derive from
manual checks by the investigators
of the system's predicted output, and it
is hard to estimate the impact of the system's suggested chunking
on the judge's determination.
We believe that the work reported here
is the first study
which has attempted to find NP chunks
subject only to the limitation
that the structures recognized
do not include recursively embedded NPs,
and which has measured performance by automatic
comparison with a preparsed corpus.

\subsection{Deriving Chunks from Treebank Parses}

We performed experiments using two different chunk structure targets,
one that tried to bracket non-recursive ``baseNPs''
and one that partitioned sentences into non-overlapping
N-type and V-type chunks,
loosely following Abney's model.
Training and test materials with chunk tags encoding
each of these kinds of structure
were derived automatically
from the parsed Wall Street Journal text
in the Penn Treebank \cite{Marcus:treebank}.
While this automatic derivation process introduced
a small percentage of errors of its own,
it was the only practical way both to provide the amount
of training data required and to allow for fully-automatic
testing.

The goal of the ``baseNP'' chunks was to identify essentially
the initial portions of non-recursive noun phrases up to the head,
including determiners but not including postmodifying
prepositional phrases or clauses.
These chunks were extracted from the Treebank parses,
basically by selecting NPs that
contained no nested NPs\footnote{This heuristic fails in some cases.
For example, Treebank uses the label NAC for some NPs
functioning as premodifiers, like ``Bank of England''
in ``Robin Leigh-Pemberton, Bank of England governor, conceded..'';
in such cases, ``governor'' is not included in any baseNP chunk.}.
The handling of conjunction followed that of the Treebank annotators
as to whether to show separate baseNPs
or a single baseNP spanning
the conjunction\footnote{Non-constituent NP conjunction,
which Treebank labels NX,
is another example that still causes problems.}.
Possessives were treated as a special case,
viewing the possessive marker
as the first word of a new baseNP,
thus flattening the recursive structure in a useful way.
The following sentences give examples of this baseNP
chunk structure:
\vspace{-18pt}
\begin{quotation} \noindent \flushleft
During \nleft~the third quarter~\nright\ ,
\nleft~Compaq~\nright\ purchased
\nleft~a former Wang Laboratories
manufacturing facility~\nright\ in
\nleft~Sterling~\nright\ ,
\nleft~Scotland~\nright~,
which will be used for
\nleft~international service and repair
operations~\nright\ .

\flushleft
\nleft~The government~\nright\ has \nleft~other agencies and
instruments~\nright\ for pursuing \nleft~these other objectives~\nright\ .

\flushleft
Even \nleft~Mao Tse-tung~\nright\ \nleft~'s China~\nright\
began in \nleft~1949~\nright\ with \nleft~a partnership~\nright\
between \nleft~the communists~\nright\ and \nleft~a number~\nright\
of \nleft~smaller , non-communist parties~\nright\ .
\end{quotation}

The chunks in the partitioning chunk experiments
were somewhat closer to Abney's model,
where the prepositions in prepositional phrases are included
with the object NP up to the head in a single N-type chunk.
This created substantial additional ambiguity for the system,
which had to distinguish prepositions from particles.
The handling of conjunction again follows the Treebank parse
with nominal conjuncts parsed in the Treebank
as a single NP forming a single N chunk,
while those parsed as conjoined NPs
become separate chunks, with any coordinating
conjunctions attached like prepositions to the following N chunk.

The portions of the text
not involved in N-type chunks
were grouped as chunks termed V-type,
though these ``V'' chunks
included many elements that were not verbal,
including adjective phrases.
The internal structure of these V-type chunks
loosely followed the Treebank parse,
though V chunks often group together elements
that were sisters in the underlying parse tree.
Again, the possessive marker was viewed as initiating a new N-type chunk.
The following sentences are annotated with these
partitioning N and V chunks:
\vspace{-18pt}
\begin{quotation} \noindent \flushleft
\nleft~Some bankers~\nright\  \vleft~are reporting~\vright\
\nleft~more inquiries than usual~\nright\
\nleft~about CDs~\nright\
\nleft~since Friday~\nright~.

\flushleft
\nleft~Eastern Airlines~\nright\
\nleft~' creditors~\nright\
\vleft~have begun exploring~\vright\
\nleft~alternative approaches~\nright\
\nleft~to a Chapter 11 reorganization~\nright\
\vleft~because~\vright\
\nleft~they~\nright
\vleft~are unhappy~\vright\
\nleft~with the carrier~\nright\
\nleft~'s latest proposal~\nright~.

\flushleft
\nleft~Indexing~\nright\ \nleft~for the most part~\nright\
\vleft~has involved simply buying~\vright\
\vleft~and then holding~\vright\ \nleft~stocks~\nright\
\nleft~in the correct mix~\nright\ \vleft~to mirror~\vright\
\nleft~a stock market barometer~\nright~.
\end{quotation}

These two kinds of chunk structure derived from the Treebank data
were encoded as chunk tags attached to each word and
provided the targets for the transformation-based learning.

\section{The Transformation-based Learning Paradigm}

\begin{figure}
\begin{center}
% for old-LaTeX use psfig, and /centering command
\epsfig{file=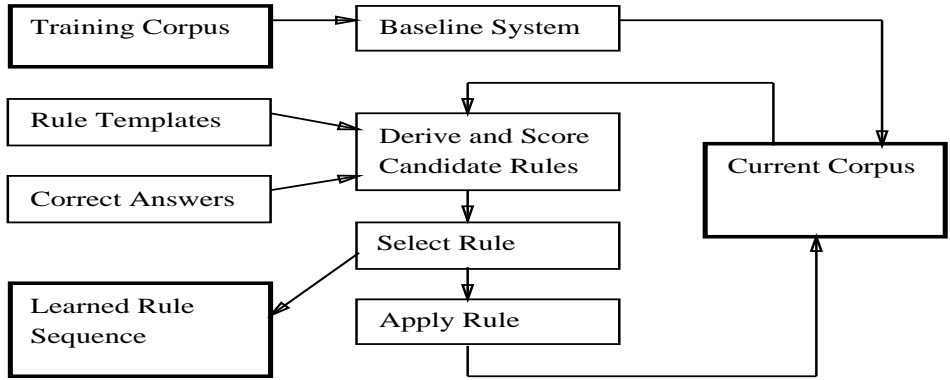,height=2in,width=5in}
\end{center}
\vspace*{-2ex}
\caption{Transformation-Based Learning}
\label{translearn}
\end{figure}

As shown in Fig.~\ref{translearn},
transformation-based learning
starts with a supervised training corpus
that specifies the correct values for some linguistic feature of interest,
a baseline heuristic for predicting initial values for that feature,
and a set of rule templates
that determine a space of possible transformational rules.
The patterns of the learned rules match to particular
combinations of features in the neighborhood surrounding a word,
and their action is to change the system's current guess
as to the feature for that word.

To learn a model,
one first applies the baseline heuristic to produce
initial hypotheses for each site in the training corpus.
At each site where this baseline prediction is not correct,
the templates are then used to form instantiated candidate rules
with patterns that test selected features in the neighborhood of the word
and actions that correct the currently incorrect tag assignment.
This process eventually identifies all the rule candidates
generated by that template set that would have a positive effect
on the current tag assignments anywhere in the corpus.

Those candidate rules are then tested against the rest of corpus,
to identify at how many locations they would cause negative changes.
One of those rules whose net score (positive changes minus negative changes)
is maximal is then
selected, applied to the corpus, and also written out as the first rule
in the learned sequence.
This entire learning process is then repeated on the transformed corpus:
deriving candidate rules, scoring them, and selecting one
with the maximal positive effect.
This process is iterated,
leading to an {\em ordered sequence} of rules,
with rules discovered first
ordered before those discovered later.
The predictions of the model on new text
are determined by beginning with the baseline heuristic prediction
and then applying each rule in the learned rule sequence in turn.

\section{Transformational Text Chunking}

This section discusses how text chunking can be encoded
as a tagging problem that can be conveniently addressed using
transformational learning.
We also note some related adaptations
in the procedure for learning rules
that improve its performance,
taking advantage of ways in which this task differs
from the learning of part-of-speech tags.

\subsection{Encoding Choices}

Applying transformational learning to text chunking requires
that the system's current hypotheses about chunk structure
be represented in a way
that can be matched against the pattern parts of rules.
One way to do this would be to have patterns match tree fragments and
actions modify tree geometries, as in Brill's
transformational parser \shortcite{Brill93a}.
In this work, we have found it convenient
to do so  by encoding the chunking
using an additional set of tags,
so that each word carries both a part-of-speech tag
and also a ``chunk tag''
from which the chunk structure can be derived.

In the baseNP experiments aimed at
non-recursive NP structures,
we use the chunk tag set
\{\tag{I}, \tag{O}, \tag{B}\}, where words marked \tag{I} are
inside some baseNP, those marked \tag{O} are outside,
and the \tag{B} tag is  used to mark
the left most item of a baseNP
which immediately follows another baseNP.
In these tests, punctuation marks were tagged
in the same way as words.

In the experiments that partitioned text into N and V chunks,
we use the chunk tag set \{\tag{BN}, \tag{N}, \tag{BV}, \tag{V}, \tag{P}\},
where \tag{BN} marks the first word and \tag{N} the succeeding words
in an N-type group while \tag{BV} and \tag{V}
play the same role for V-type groups.
Punctuation marks,
which are ignored in Abney's chunk grammar,
but which the Treebank data treats as normal lexical items
with their own part-of-speech tags,
are unambiguously assigned the chunk tag \tag{P}.
Items tagged \tag{P} are allowed to appear within N or V chunks;
they are irrelevant as far as chunk boundaries are concerned,
but they are still available to be matched against
as elements of the left hand sides of rules.

Encoding chunk structure
with tags attached to words
rather than non-recursive bracket markers inserted
between words
has the advantage
that it limits the dependence
between different elements of the encoded representation.
While brackets must be correctly paired
in order to derive a chunk structure,
it is easy to define a mapping
that can produce a valid chunk structure
from any sequence of chunk tags;
the few hard cases that arise
can be handled completely locally.
For example, in the baseNP tag set,
whenever a \tag{B} tag immediately follows an \tag{O}, it
must be treated as an \tag{I},
and, in the partitioning chunk tag set,
wherever a \tag{V} tag immediately follows an \tag{N} tag
without any intervening \tag{BV},
it must be treated as a \tag{BV}.

\subsection{Baseline System}

Transformational learning begins with some initial ``baseline''
prediction, which here means
a baseline assignment of chunk tags to words.
Reasonable suggestions for baseline heuristics
after a text has been tagged for part-of-speech
might
include assigning to each {\em word} the chunk tag that it carried
most frequently in the training set,
or assigning each {\em part-of-speech tag}
the chunk tag that was most frequently associated
with that part-of-speech tag in the training.
We tested both approaches, and the baseline heuristic
using part-of-speech tags turned out to do better,
%as one might expect,
so it was the one used in our experiments.
The part-of-speech tags used by this baseline heuristic,
and then later also matched against by transformational rule patterns,
were derived by running the raw texts in a prepass through
Brill's transformational part-of-speech tagger \cite{Brill:tagger}.

\subsection{Rule Templates}

In transformational learning, the space of candidate rules
to be searched is defined by a set of rule templates
that each specify a small number of particular feature sets
as the relevant factors that a rule's
left-hand-side pattern should examine,
for example, the part-of-speech tag
of the word two to the left
combined with the actual word one to the left.
In the preliminary scan of the corpus for each learning pass,
it is these templates that are applied
to each location whose current tag is not correct,
generating a candidate rule that would apply at least at that one location,
matching those factors and correcting the chunk tag assignment.

When this approach is applied to part-of-speech tagging,
the possible sources of evidence for templates
involve the identities of words
within a neighborhood of some appropriate size
and their current part-of-speech tag assignments.
In the text chunking application,
the tags being assigned are chunk structure tags,
while the part-of-speech tags are a fixed part of the environment,
like the lexical identities of the words themselves.
This additional class of available information causes a significant
increase in the number of reasonable templates
if templates for a wide range of the possible combinations of evidence
are desired.
The distributed version of Brill's tagger \cite{Brill:tagger}
makes use of 26 templates, involving
various mixes of word and part-of-speech tests on neighboring words.
Our tests were performed using 100 templates;
these included almost all of Brill's combinations,
and extended them to include references to chunk tags
as well as to words and part-of-speech tags.

The set of 100 rule templates used here
was built from repetitions of 10 basic patterns, shown
on the left side of
Table~\ref{word:patterns} as they apply to words.
\begin{table}[b]
\centering
\begin{tabular}{|l|l|l|l|}
\hline
\multicolumn{2}{|c|}{Word Patterns}&\multicolumn{2}{|c|}{Tag Patterns}\\
\hline
Pattern & Meaning & Pattern & Meaning \\
\hline
\pat{W}{0} & current word & \pat{T}{0} &
  current tag \\
\pat{W}{-1} & word 1 to left & \pat{T}{-1}, \pat{T}{0} &
  current tag and tag to left \\
\pat{W}{1} & word 1 to right & \pat{T}{0}, \pat{T}{1} &
  current tag and tag to right \\
\pat{W}{-1}, \pat{W}{0} & current word and word to left &
  \pat{T}{-2}, \pat{T}{-1} & two tags to left \\
\pat{W}{0}, \pat{W}{1} & current word and word to right &
  \pat{T}{1}, \pat{T}{2} & two tags to right \\
\pat{W}{-1}, \pat{W}{1} & word to left and word to right & &\\
\pat{W}{-2}, \pat{W}{-1} & two words to left & & \\
\pat{W}{1}, \pat{W}{2} & two words to right & & \\
\pat{W}{-1,-2,-3} & word 1 or 2 or 3 to left & & \\
\pat{W}{1,2,3} & word 1 or 2 or 3 to right & & \\
\hline
\end{tabular}
\caption{Patterns used in Templates}
\label{word:patterns}
\end{table}
The same 10 patterns can also be used to match against
part-of-speech tags, encoded as \pat{P}{0}, \pat{P}{-1}, etc.
(In other tests, we have explored mixed templates,
that match against both word and part-of-speech values,
but no mixed templates were used in these experiments.)
These 20 word and part-of-speech patterns were then combined with
each of the 5 different chunk tag patterns
shown on the right side of the table.
The cross product of the 20 word and part-of-speech patterns
with the 5 chunk tag patterns
determined the full set of 100 templates used.

\section{Algorithm Design Issues}

The large increase in the number of rule templates
in the text chunking application
when compared to part-of-speech tagging
pushed the training process against the available limits
in terms of both space and time,
particularly when combined with the desire
to work with the largest possible training sets.
Various optimizations proved to be crucial
to make the tests described feasible.

\subsection{Organization of the Computation}

One change in the algorithm is related
to the smaller size of the tag set.
In Brill's tagger \cite{Brill:tagger},
an initial calculation in each pass computes the confusion matrix
for the current tag assignments and
sorts the entries of that [old-tag $\times$ new-tag] matrix,
so that candidate rules can then be processed
in decreasing order of the maximum possible benefit
for any rule changing, say, old tag I to new tag J.
The search for the best-scoring rule can then be halted
when a cell of the confusion matrix is reached
whose maximum possible benefit is less
than the net benefit of some rule already encountered.

The power of that approach is dependent on the fact
that the confusion matrix for part-of-speech tagging
partitions the space of candidate rules
into a relatively large number of classes,
so that one is likely to be able to exclude
a reasonably large portion of the search space.
In a chunk tagging application, with only 3 or 4 tags
in the effective tagset,
this approach based on the confusion matrix offers much less benefit.

However, even though the confusion matrix does not usefully subdivide
the space of possible rules when the tag set is this small,
it is still possible to apply a similar optimization by sorting the entire list
of candidate rules on the basis of their positive scores,
and then processing the candidate rules
(which means determining their negative scores and thus their net scores)
in order of decreasing positive scores.
By keeping track of the rule with maximum benefit seen so far,
one can be certain of having found one of the globally best rules
when one reaches candidate rules in the sorted list
whose positive score is
not greater than the net score of the best rule so far.

\subsection{Indexing Static Rule Elements}

In earlier work on transformational part-of-speech
tagging \cite{RamshawMarcus94},
we noted that it is possible to greatly speed up the learning process
by constructing a full, bidirectional index
linking each candidate rule to those locations in the corpus
at which it applies
and each location in the corpus to those candidate rules
that apply there.
Such an index allows the process of applying rules
to be performed without having to search through the corpus.
Unfortunately, such complete indexing proved to be
too costly in terms of physical memory to be feasible
in this application.

However, it is possible to construct a limited index
that lists for each candidate rule those locations in the corpus
at which the {\em static} portions of its left-hand-side pattern match.
Because this index involves only the stable word identity and
part-of-speech tag values, it does not require updating;
thus it can be stored more compactly, and
it is also not necessary to maintain back pointers
from corpus locations to the applicable rules.
This kind of partial static index proved to be a significant advantage
in the portion of the program where candidate rules
with relatively high positive scores are being tested
to determine their negative scores,
since it avoids the necessity of testing such rules against
every location in the corpus.

\subsection{Heuristic Disabling of Unlikely Rules}

We also investigated a new heuristic to
speed up the computation:
After each pass, we disable
all rules whose positive score is significantly lower
than the net score of the best rule for the current pass.
A disabled rule is then reenabled whenever enough other changes
have been made to the corpus that it seems possible that the score
of that rule might have changed enough to bring it back into contention
for the top place.
This is done by adding some fraction of the changes made in each pass
to the positive scores of the disabled rules,
and reenabling rules whose adjusted positive scores came within a threshold
of the net score of the successful rule on some pass.

Note that this heuristic technique introduces some risk of missing
the actual best rule in a pass, due to its being incorrectly disabled
at the time.
However, empirical comparisons between runs with and without rule disabling
suggest that conservative use of this technique
can produce an order of magnitude speedup
while imposing only a very slight cost in terms of suboptimality
of the resulting learned rule sequence.

\section{Results}

The automatic derivation of training and testing data from
the Treebank analyses
allowed for fully automatic scoring,
though the scores are naturally subject to any remaining
systematic errors in the data derivation process
as well as to {\em bona fide} parsing errors
in the Treebank source.
Table~\ref{basenp:results} shows the results
for the baseNP tests, and
Table~\ref{chunk:results} shows the results
for the partitioning chunks task.
Since training set size has a significant effect
on the results,
values are shown for three different training set sizes.
(The test set in all cases was 50K words.
Training runs were halted after the first 500 rules;
rules learned after that point affect relatively few locations
in the training set and have only a very slight effect
for good or ill on test set performance.)

\begin{table}
\centering
\begin{tabular}{|c|cc|cc|cc|}
\hline
Training & Recall & Error Red. & Precision & Error Red. & Corr. Tag & Error
Red. \\
\hline
Baseline & 81.9\% & & 78.2\% & & 94.5\% & \\
 50K & 90.4\% & 47.2\% & 89.8\% & 53.1\% & 96.9\% & 44.4\% \\
100K & 91.8\% & 54.8\% & 91.3\% & 60.0\% & 97.2\% & 49.6\% \\
200K & 92.3\% & 57.4\% & 91.8\% & 62.4\% & 97.4\% & 53.4\% \\
\hline
\end{tabular}
\caption{BaseNP Chunk Results}
\label{basenp:results}
\end{table}

\begin{table}
\centering
\begin{tabular}{|c|cc|cc|cc|}
\hline
Training & Recall & Error Red. & Precision & Error Red. & Corr. Tag & Error
Red. \\
\hline
Baseline & 60.0\% & & 47.8\% & & 78.0\% & \\
 50K & 86.6\% & 66.6\% & 85.8\% & 72.8\% & 94.4\% & 74.4\% \\
100K & 88.2\% & 70.4\% & 87.4\% & 75.8\% & 95.0\% & 77.3\% \\
200K & 88.5\% & 71.1\% & 87.7\% & 76.5\% & 95.3\% & 78.5\% \\
\hline
\end{tabular}
\caption{Partitioning Chunk Results}
\label{chunk:results}
\end{table}

The first line in each table gives the performance of the baseline system,
which assigned a baseNP or chunk tag to each word
on the basis of the POS tag assigned in the prepass.
Performance is stated in terms of recall
(percentage of correct chunks found)
and precision
(percentage of chunks found that are correct),
where both ends of a chunk had to match exactly for it to be counted.
The raw percentage of correct chunk tags is also given
for each run, and for each performance measure,
the relative error reduction compared to the baseline is listed.
The partitioning chunks do appear
to be somewhat harder to predict than baseNP chunks.
The higher error reduction for the former
is partly due to the fact that the part-of-speech baseline
for that task is much lower.

\subsection{Analysis of Initial Rules}

To give a sense of the kinds of rules being learned,
the first 10 rules from the 200K baseNP run
are shown in Table~\ref{basenp:examples}.
It is worth glossing the rules,
since one of the advantages of transformation-based learning is
exactly that the resulting model is easily interpretable.
In the first of the baseNP rules,
adjectives (with part-of-speech tag JJ)
that are currently tagged \tag{I}
but that are followed by words tagged \tag{O}
have their tags changed to \tag{O}.
In Rule~2, determiners that are preceded by two words
both tagged \tag{I} have their own tag changed to \tag{B},
marking the beginning of a baseNP
that happens to directly follow another.
(Since the tag \tag{B} is only used when baseNPs abut,
the baseline system tags determiners as \tag{I}.)
Rule~3 takes words which immediately follow
determiners
tagged \tag{I} that in turn follow something tagged \tag{O}
and changes their tag to also be \tag{I}.
Rules 4--6 are similar to Rule~2, marking the initial words
of baseNPs that directly follow another baseNP.
Rule~7 marks conjunctions (with part-of-speech tag CC) as \tag{I}
if they follow an \tag{I} and precede a noun,
since such conjunctions are more likely to be embedded in a single baseNP
than to separate two baseNPs,
and Rules 8 and 9 do the same.
(The word ``\&'' in rule~8 comes mostly from company names
in the Wall St. Journal source data.)
Finally, Rule~10 picks up cases like ``including about four million shares''
where ``about'' is used
as a quantifier rather than preposition.

\begin{table}[th!]
\centering
\begin{tabular}{|r|l|l|l|}
\hline
Pass & Old Tag & Context & New Tag \\
\hline
1. & \tagi{I} & \slot{T}{1}{\tag{O}}, \slot{P}{0}{JJ} & \tagi{O} \\
2. & \tagi{-} & \slot{T}{-2}{\tag{I}}, \slot{T}{-1}{\tag{I}}, \slot{P}{0}{DT} &
\tagi{B} \\
3. & \tagi{-} & \slot{T}{-2}{\tag{O}}, \slot{T}{-1}{I}, \slot{P}{-1}{DT} &
\tagi{I} \\
4. & \tagi{I} & \slot{T}{-1}{\tag{I}}, \slot{P}{0}{WDT} & \tagi{B} \\
5. & \tagi{I} & \slot{T}{-1}{\tag{I}}, \slot{P}{0}{PRP} & \tagi{B} \\
6. & \tagi{I} & \slot{T}{-1}{\tag{I}}, \slot{W}{0}{who} & \tagi{B} \\
7. & \tagi{O} & \slot{T}{-1}{\tag{I}}, \slot{P}{0}{CC}, \slot{P}{1}{NN} &
\tagi{I} \\
8. & \tagi{O} & \slot{T}{1}{\tag{I}}, \slot{W}{0}{\&} & \tagi{I} \\
9. & \tagi{O} & \slot{T}{-1}{\tag{I}}, \slot{P}{0}{CC}, \slot{P}{1}{NNS} &
\tagi{I} \\
10. & \tagi{O} & \slot{T}{-1}{\tag{O}}, \slot{W}{0}{about} & \tagi{I} \\
\hline
\end{tabular}
\caption{First Ten Basenp Chunk Rules}
\label{basenp:examples}
\end{table}

A similar list of the first ten rules for the chunk task
can be seen in Table~\ref{chunking:examples}.
To gloss a few of these,
in the first rule here, determiners (with part-of-speech tag DT),
which usually begin N chunks and thus are assigned the baseline
tag \tag{BN}, have their chunk tags
changed to \tag{N} if they follow a word whose tag is also \tag{BN}.
In Rule~2, sites currently tagged \tag{N} but which fall
at the beginning of a sentence have their tags switched to \tag{BN}.
(The dummy tag \tag{Z} and word ZZZ
indicate that the locations one to the left are beyond the
sentence boundaries.)
Rule~3 changes \tag{N} to \tag{BN} after a comma (which is tagged \tag{P}), and
in Rule~4, locations tagged \tag{BN} are switched to \tag{BV} if the following
location is tagged \tag{V} and has the part-of-speech tag VB.

\begin{table}[th!]
\centering
\begin{tabular}{|r|l|l|l|}
\hline
Pass & Old Tag & Context & New Tag \\
\hline
1. & \tagi{BN} & \slot{T}{-1}{\tag{BN}}, \slot{P}{0}{DT} & \tagi{N} \\
2. & \tagi{N} & \slot{T}{-1}{\tag{Z}}, \slot{W}{-1}{ZZZ} & \tagi{BN} \\
3. & \tagi{N} & \slot{T}{-1}{\tag{P}}, \slot{P}{-1}{`,'} & \tagi{BN} \\
4. & \tagi{BN} & \slot{T}{1}{\tag{V}}, \slot{P}{1}{VB} & \tagi{BV} \\
5. & \tagi{N} & \slot{T}{-1}{\tag{BV}}, \slot{P}{-1,-2,-3}{VBD} & \tagi{BN} \\
6. & \tagi{N} & \slot{P}{-1}{VB} & \tagi{BN} \\
7. & \tagi{BV} & \slot{T}{-1}{\tag{V}}, \slot{P}{-1,-2,-3}{RB} & \tagi{V} \\
8. & \tagi{V} & \slot{T}{-1}{\tag{N}}, \slot{P}{-1,-2,-3}{NN} & \tagi{BV} \\
9. & \tagi{BV} & \slot{T}{-1}{\tag{BV}}, \slot{P}{1,2,3}{VB} & \tagi{V} \\
10. & \tagi{BN} & \slot{T}{-1}{\tag{BN}}, \slot{P}{0}{PRP\$} & \tagi{N} \\
\hline
\end{tabular}
\caption{First Ten Partitioning Chunk Rules}
\label{chunking:examples}
\end{table}

\subsection{Contribution of Lexical Templates}

The fact that this system includes lexical rule templates
that refer to actual words sets it apart from approaches
that rely only on part-of-speech tags to predict chunk structure.
To explore how much difference in performance those lexical rule
templates make, we repeated the above test runs omitting
templates that refer to specific words.
The results for these runs, in Tables~\ref{basenp:nolex:results}
and \ref{chunk:nolex:results},
suggest that the lexical rules improve performance
on the baseNP chunk task by about 1\%
(roughly 5\% of the overall error reduction)
and on the partitioning chunk task by about 5\%
(roughly 10\% of the error reduction).
Thus lexical rules appear to be making a limited contribution
in determining baseNP chunks,
but a more significant one for the partitioning chunks.

\begin{table}[th!]
\centering
\begin{tabular}{|c|cc|cc|cc|}
\hline
Training & Recall & Error Red. & Precision & Error Red. & Corr. Tag & Error
Red. \\
\hline
Baseline & 81.9\% & & 78.2\% & & 94.5\% & \\
 50K & 89.6\% & 42.7\% & 88.9\% & 49.2\% & 96.6\% & 38.8\% \\
100K & 90.6\% & 48.4\% & 89.9\% & 53.7\% & 96.9\% & 44.4\% \\
200K & 90.7\% & 48.7\% & 90.5\% & 56.3\% & 97.0\% & 46.0\% \\
\hline
\end{tabular}
\caption{BaseNP Chunk Results Without Lexical Templates}
\label{basenp:nolex:results}
\end{table}

\begin{table}[th!]
\centering
\begin{tabular}{|c|cc|cc|cc|}
\hline
Training & Recall & Error Red. & Precision & Error Red. & Corr. Tag & Error
Red. \\
\hline
Baseline & 60.0\% & & 47.8\% & & 78.0\% & \\
 50K & 81.8\% & 54.5\% & 81.4\% & 64.4\% & 92.4\% & 65.4\% \\
100K & 82.9\% & 57.2\% & 83.0\% & 67.3\% & 92.9\% & 67.9\% \\
200K & 83.6\% & 58.9\% & 83.5\% & 68.4\% & 93.9\% & 72.2\% \\
\hline
\end{tabular}
\caption{Partitioning Chunk Results Without Lexical Templates}
\label{chunk:nolex:results}
\end{table}

\subsection{Frequent Error Classes}

A rough hand categorization of a sample of the errors from a baseNP run
indicates that many fall into
classes that are understandably difficult for any process
using only local word and part-of-speech patterns to resolve.
The most frequent single confusion involved words tagged VBG and VBN,
whose baseline prediction given their part-of-speech tag
was \tag{O}, but which also occur frequently inside baseNPs.
The system did discover some rules that allowed it
to fix certain classes of VBG and VBN mistaggings,
for example, rules that retagged VBNs as \tag{I}
when they preceded an NN or NNS tagged \tag{I}.
However, many also remained unresolved, and many of those
appear to be cases that would require more than local word and
part-of-speech patterns to resolve.

The second most common class of errors involved conjunctions,
which, combined with the former class,
make up half of all the errors in the sample.
The Treebank tags the words ``and'' and frequently ``,''
with the part-of-speech tag CC,
which the baseline system again predicted would fall most often
outside of a baseNP\footnote{Note that this is
one of the cases where Church's chunker
allows separate NP fragments to count as chunks.}.
However, the Treebank parses do also frequently classify
conjunctions of Ns or NPs as a single baseNP,
and again there appear to be insufficient clues in the word and tag
contexts for the current system to make the distinction.
Frequently, in fact, the actual choice of structure
assigned by the Treebank annotators seemed largely dependent on
semantic indications unavailable to the transformational learner.

\section{Future Directions}

We are planning to explore several different paths
that might increase the system's power to distinguish
the linguistic contexts in which particular changes would be useful.
One such direction is to expand the template set
by adding templates that are sensitive to the chunk structure.
For example, instead of referring to the word two to the left,
a rule pattern could refer to the first word in the current chunk,
or the last word of the previous chunk.
Another direction would be to enrich the vocabulary of chunk tags,
so that they could be used during the learning process to encode
contextual features for use by later rules in the sequence.

We would also like to explore applying these same kinds of techniques
to building
larger scale structures, in which larger units are assembled
or predicate/argument structures derived by combining chunks.
One interesting direction here would be to explore the use of
chunk structure tags that encode a form of dependency grammar,
where the tag ``N+2'' might mean that the current word is to be taken
as part of the unit headed by the N two words to the right.

\section{Conclusions}

% ``kind of'' added to avoid a bad line break on a hyphenated word
By representing text chunking as a kind of tagging problem,
it becomes possible to easily apply transformation-based learning.
We have shown that this approach
is able to automatically induce
a chunking model from supervised training
that achieves recall and precision of 92\% for baseNP chunks
and 88\% for partitioning N and V chunks.
Such chunking models provide a useful and feasible next step
in textual interpretation that goes beyond part-of-speech tagging,
and that serve as a foundation both for larger-scale grouping
and for direct extraction of subunits like index terms.
In addition, some variations in the transformation-based learning algorithm
are suggested by this application that may also be useful in other settings.

\section*{Acknowledgments}

We would like to thank Eric Brill for making his system widely available,
and Ted Briscoe and David Yarowsky for helpful
comments, including the suggestion to test the system's
performance without lexical rule templates.


\begin{thebibliography}{}

\bibitem[\protect\citename{Abney}1991]{Abney91}
Abney, Steven.
\newblock 1991.
\newblock Parsing by chunks.
\newblock In Berwick, Abney, and Tenny, editors, {\em Principle-Based Parsing}.
  Kluwer Academic Publishers.

\bibitem[\protect\citename{Bourigault}1992]{Bourigault92}
Bourigault, D.
\newblock 1992.
\newblock Surface grammatical analysis for the extraction of terminological
  noun phrases.
\newblock In {\em Proceedings of the Fifteenth International Conference on
  Computational Linguistics}, pages 977--981.

\bibitem[\protect\citename{Brill}1993a]{Brill93a}
Brill, Eric.
\newblock 1993a.
\newblock Automatic grammar induction and parsing free text: A
  transformation-based approach.
\newblock In {\em Proceedings of the DARPA Speech and Natural Language
  Workshop, 1993}, pages 237--242.

\bibitem[\protect\citename{Brill}1993b]{Brill93}
Brill, Eric.
\newblock 1993b.
\newblock {\em A Corpus-Based Approach to Language Learning}.
\newblock {Ph.D.} thesis, University of Pennsylvania.

\bibitem[\protect\citename{Brill}1993c]{Brill:tagger}
Brill, Eric.
\newblock 1993c.
\newblock Rule based tagger, version 1.14.
\newblock Available from ftp.cs.jhu.edu in the directory /pub/brill/programs/.

\bibitem[\protect\citename{Brill}1994]{Brill94b}
Brill, Eric.
\newblock 1994.
\newblock Some advances in transformation-based part of speech tagging.
\newblock In {\em Proceedings of the Twelfth National Conference on Artificial
  Intelligence}, pages 722--727.
\newblock (cmp-lg/9406010).

\bibitem[\protect\citename{Brill and Resnik}1994]{BrillResnik94}
Brill, Eric and Philip Resnik.
\newblock 1994.
\newblock A rule-based approach to prepositional attachment disambiguation.
\newblock In {\em Proceedings of the Sixteenth International Conference on
  Computational Linguistics}.
\newblock (cmp-lg/9410026).

\bibitem[\protect\citename{Church}1988]{Church88}
Church, Kenneth.
\newblock 1988.
\newblock A stochastic parts program and noun phrase parser for unrestricted
  text.
\newblock In {\em Second Conference on Applied Natural Language Processing}.
  ACL.

\bibitem[\protect\citename{Ejerhed}1988]{Ejerhed88}
Ejerhed, Eva~I.
\newblock 1988.
\newblock Finding clauses in unrestricted text by finitary and stochastic
  methods.
\newblock In {\em Second Conference on Applied Natural Language Processing},
  pages 219--227. ACL.

\bibitem[\protect\citename{Gee and Grosjean}1983]{GeeGrosjean83}
Gee, James~Paul and Fran\c{c}ois Grosjean.
\newblock 1983.
\newblock Performance structures: A psycholinguistic and linguistic appraisal.
\newblock {\em Cognitive Psychology}, 15:411--458.

\bibitem[\protect\citename{Kupiec}1993]{Kupiec93}
Kupiec, Julian.
\newblock 1993.
\newblock An algorithm for finding noun phrase correspondences in bilingual
  corpora.
\newblock In {\em Proceedings of the 31st Annual Meeting of the ACL}, pages
  17--22.

\bibitem[\protect\citename{Marcus \bgroup et al.\egroup }1994]{Marcus:treebank}
Marcus, Mitchell, Grace Kim, Mary~Ann Marcinkiewicz, Robert MacIntyre, Ann
  Bies, Mark Ferguson, Karen Katz, and Britta Schasberger.
\newblock 1994.
\newblock The {P}enn {T}reebank: A revised corpus design for extracting
  predicate argument structure.
\newblock In {\em Human Language Technology, ARPA March 1994 Workshop}. Morgan
  Kaumann.

\bibitem[\protect\citename{Ramshaw and Marcus}1994]{RamshawMarcus94}
Ramshaw, Lance~A. and Mitchell~P. Marcus.
\newblock 1994.
\newblock Exploring the statistical derivation of transformational rule
  sequences for part-of-speech tagging.
\newblock In {\em Proceedings of the ACL Balancing Act Workshop on Combining
  Symbolic and Statistical Approaches to Language}, pages 86--95.
\newblock (cmp-lg/9406011).

\bibitem[\protect\citename{Voutilainen}1993]{Voutilainen93}
Voutilainen, Atro.
\newblock 1993.
\newblock {NPT}ool, a detector of {E}nglish noun phrases.
\newblock In {\em Proceedings of the Workshop on Very Large Corpora}, pages
  48--57. ACL, June.
\newblock (cmp-lg/9502010).

\end{thebibliography}
\end{document}